\documentclass{article}

\usepackage{PRIMEarxiv}

\usepackage[utf8]{inputenc} 
\usepackage[T1]{fontenc}    
\usepackage{hyperref}       
\usepackage{url}            
\usepackage{booktabs}       
\usepackage{amsfonts}       
\usepackage{nicefrac}       

\usepackage{multirow}
\usepackage{amsmath}

\usepackage{microtype}      
\usepackage{lipsum}
\usepackage{fancyhdr}       
\usepackage{graphicx,comment} 
\usepackage{hyperref}       
\graphicspath{{media/}}     

\title{3D-CBM: A Framework for Concept-Based Interpretability in Generative 3D Modeling
}

\author{
  Ahmad Al-Kabbany \\
  Yubree Labs, Ottawa K2M1T2, Canada \\
  Multimedia Interaction and Communication Lab, Arab Academy for Science and Technology, Alexandria 21937, Egypt  \\
  \texttt{alkabbany@ieee.org, alkabbany@aast.edu} 
}

\begin{document}
\maketitle

\begin{abstract}
This research introduces a framework for incorporating Concept Bottleneck Models (CBMs) into 3D generative architectures to address the inherent 'semantic gap' in deep geometric learning. As deep models become central to 3D content creation, explainability shifts from a peripheral feature to a fundamental requirement for trust and accountability in safety-critical domains such as manufacturing and healthcare. CBMs provide an intrinsic interpretability solution by constraining latent representations to align with human-defined concepts, yet their application to unstructured 3D data remains largely unexplored.

We design, implement, and validate a formal 3D-CBM architecture that maps raw geometric inputs, including point clouds and meshes, into a multi-tiered taxonomy of interpretable primitives and functional attributes. The framework further identifies strategic datasets, such as PartNet and ShapeNet, specialized for concept-based supervision. Results of a 3D part-manipulation proof-of-concept experiment demonstrate the framework's efficacy, achieving a concept prediction accuracy of 88.8\% and a Chamfer Distance of 0.0115. Critically, the model enables precise test-time intervention, allowing for the interactive correction of structural errors. This work establishes a foundation for semantically-steerable 3D generation and invites further exploration into collaborative human-in-the-loop design systems.
\end{abstract}

\keywords{3D Generative Models \and Concept Bottleneck Models \and Explainable AI \and Human-in-the-Loop \and Point Clouds \and 3D Mesh Processing \and Semantic Steerability \and Geometric Deep Learning \and Trustworthy AI \and Digital Content Creation}

\section{Introduction}
While modern generative architectures can produce 3D meshes with millions of polygons and stunning geometric detail, a profound "semantic gap" remains: we can generate a complex object, yet we cannot easily explain to the model that the legs of a generated table are too thin to support its weight or that a structural joint is anatomically incorrect \cite{po2024compositional,po2024state}. This disconnect highlights the urgent need for Explainable AI (XAI) to serve as a bridge between high-dimensional model complexity and human trust. Within this landscape, Concept Bottleneck Models (CBMs) represent a fundamental paradigm shift from post-hoc interpretation toward intrinsic transparency\cite{yuksekgonul2022post}. Rather than attempting to decode a "black box" after a decision is made, CBMs are engineered as "glass boxes" by design. They achieve this by constraining internal latent representations to pass through a predefined layer of human-understandable concepts. By transforming abstract features into a structured vocabulary of discrete, interpretable attributes—such as "planarity," "symmetry," or "part-thickness"—CBMs provide a framework where the path from input to 3D output is not only observable but logically aligned with human reasoning and physical constraints.

The fundamental distinction between Concept Bottleneck Models and classical XAI techniques lies in the transition from post-hoc observation to intrinsic interpretability. Dominant 3D explainability methods—such as saliency maps or Grad-CAM—are primarily diagnostic; they highlight where a model focuses its attention, but the semantic meaning of those highlighted regions remains subject to user interpretation \cite{ahmadi2024explainability}. In contrast, CBMs are designed to reveal what the model has actually understood by decomposing its internal logic into human-defined concepts. This architectural shift has profound implications for 3D modeling applications. While post-hoc methods are often unfaithful to the model's true reasoning or sensitive to input noise, CBMs ensure that the explanation is inherently tied to the decision-making process. Perhaps most critically, CBMs enable \textbf{test-time intervention}, a capability entirely absent in standard XAI. In a generative context, this allows a designer to interactively rectify a model’s mistakes—such as manually adjusting a "structural stability" concept—to steer the 3D output toward a more accurate and physically viable result.

The efficacy of CBMs has been extensively demonstrated across diverse 2D computer vision domains, where they have consistently bridged the gap between model performance and human trust. In safety-critical fields such as medical imaging, CBMs have transformed diagnostic workflows by predicting human-interpretable clinical concepts—such as cell nucleus shapes or visual cues in chest X-rays and CT—prior to making a final classification \cite{oikarinen2023label,khaled2026interpretable}. These models have shown that incorporating clinical knowledge can actually enhance performance on out-of-domain data, proving that interpretability does not necessarily come at the expense of accuracy. Beyond healthcare, CBMs have significantly impacted fine-grained visual recognition tasks, such as bird species classification, by anchoring predictions in discrete visual attributes like "wing color" or "beak shape". Recent advancements have even scaled this paradigm to massive vision-language models and generative tasks in natural language processing and protein design, where "interpretable neurons" \cite{yang2023language,dang2024explainable} allow for safer, more controlled generation. The proven ability of CBMs to enable precise concept detection and high-level steerability in these 2D and 1D applications suggests a transformative potential for the increasingly complex world of 3D deep learning.

Despite the clear advantages of concept-based interpretability, its application to the 3D domain remains largely unexplored, leaving a significant gap in the field of 3D deep learning \cite{khaled2026interpretable}. Current 3D XAI research is primarily dominated by post-hoc diagnostic tools that struggle with the unique complexities of 3D data representations, such as the unstructured nature of point clouds and the varying topologies of meshes. Methods that rely on heat-maps or geometric perturbations often yield "noisy" explanations that are difficult for designers to translate into actionable insights, as they highlight raw geometric features rather than semantic intent. This lack of transparency is particularly problematic given the recent explosion in 3D generative AI, where models can now produce complex assets in seconds but offer users little to no control over the internal structural logic \cite{li2023generative,jiang2024survey}. The "semantic gap" previously mentioned—where a model might generate a visually plausible object that is structurally unsound—cannot be bridged by post-hoc methods alone. Consequently, the introduction of CBMs into 3D modeling is not merely an incremental improvement but a necessary evolution; by anchoring generation in a "glass-box" layer of human-defined geometric and functional concepts, we can finally move beyond passive observation toward a paradigm of authoritative, steerable, and trustworthy 3D creation.

\begin{figure*}[t]
    \centering
    \includegraphics[width=\textwidth]{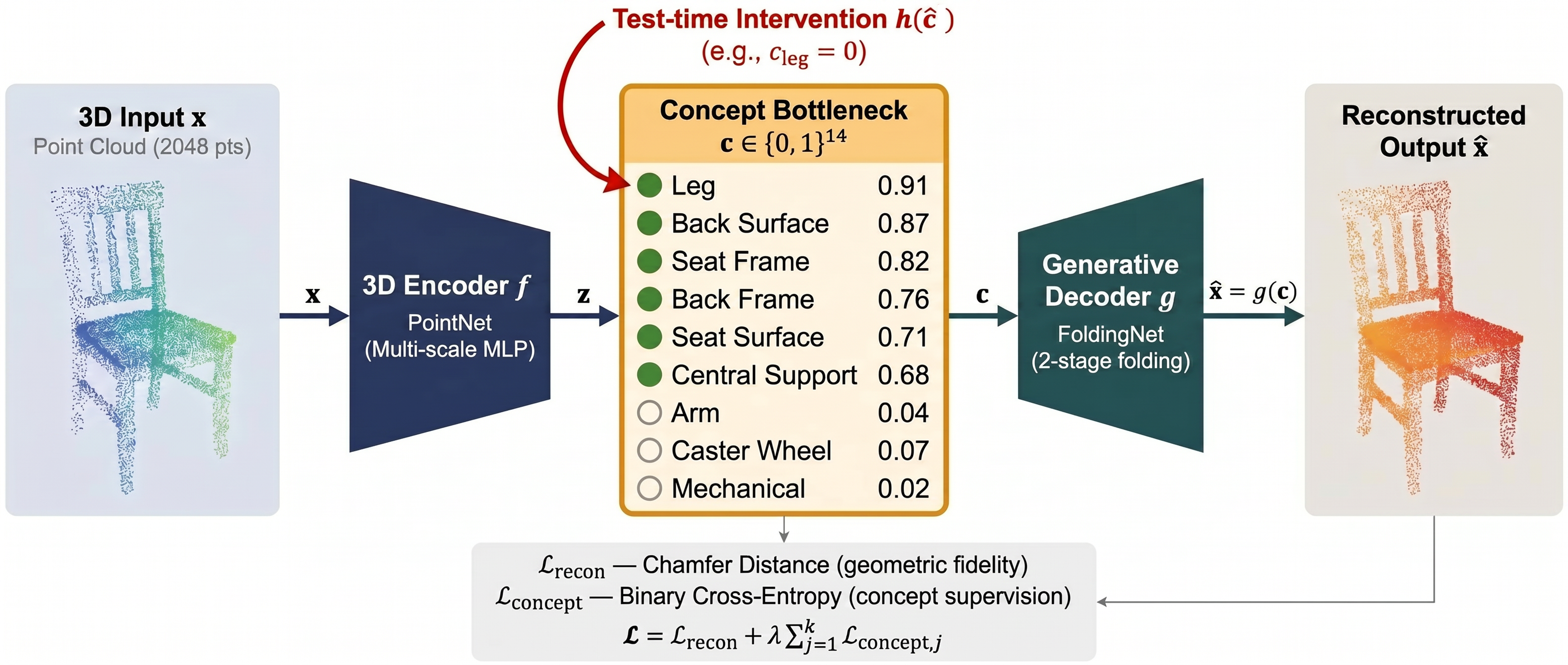}
    \caption{The proposed 3D-CBM pipeline. A 3D point cloud is encoded
    into a human-interpretable Concept Bottleneck layer before being
    decoded into a reconstructed point cloud. The bottleneck enables
    test-time intervention, allowing a domain expert to directly override
    individual concept activations --- such as forcing $c_{\text{leg}} = 0$
    --- to steer the structural properties of the generative output.}
    \label{fig:pipeline}
\end{figure*}

To address these critical limitations, this paper introduces a novel framework for incorporating Concept Bottleneck Models into 3D generative architectures, marking a significant step toward semantically steerable 3D deep learning. Our contribution moves beyond traditional 3D XAI by embedding human-defined geometric and structural concepts directly into the generative pipeline, thereby enabling both intrinsic transparency and post-generation intervention. By formalizing the mapping from unstructured 3D representations—such as point clouds and meshes—to a structured concept bottleneck, we establish a foundation for models that are not only interpretable but also physically and semantically consistent with human design principles. We posit that this approach provides the necessary "glass-box" mechanism for high-stakes 3D applications, from automated CAD design to medical reconstruction.The proposed 3D-CBM pipeline is depicted in Fig.~\ref{fig:pipeline}. Specifically, this work makes the following contributions:
\begin{enumerate}
  \item \textbf{A Formal 3D-CBM Mathematical Framework}: We define a generalized architecture for mapping non-Euclidean 3D data into a human-interpretable concept layer, supporting both discriminative and generative tasks.
  
  \item \textbf{3D Concept Taxonomy and Hierarchy}: We propose a multi-tiered taxonomy of 3D-specific concepts, categorizing them into low-level geometric primitives (e.g., planarity, curvature), mid-level structural parts, and high-level functional attributes (e.g., structural stability).
  
  \item \textbf{Protocol for Human-in-the-Loop 3D Intervention}: We discuss and explain a methodology for test-time intervention, allowing users to interactively modify concept activations to rectify geometric errors and steer model output.
  
  \item \textbf{Comprehensive Mapping of 3D Datasets for CBM Training}: We provide a review and categorization of existing 3D datasets, such as PartNet and ShapeNet, that are uniquely suited for training concept-based models.
\end{enumerate}

The remainder of this article is organized to provide an exploration of the 3D-CBM landscape. Section 2 presents a detailed literature review, contextualizing our work within the broader evolution of 3D deep learning and existing XAI paradigms. Section 3 details the core methodology of the proposed framework, formalizing the 3D Concept Bottleneck architecture and defining the multi-tiered concept taxonomy. In Section 4, we discuss the strategic selection of 3D datasets and their adaptation for concept-based supervision. Section 5 presents a "Proof-of-Concept" experiment, demonstrating the framework's viability in extracting and manipulating geometric concepts within a controlled 3D environment. Finally, Section 6 concludes the study with a vision for future research in semantically-aware generative modeling.

\section{Literature Review}
\label{sec:literature}

The development of the proposed 3D Concept Bottleneck Model (3D-CBM) framework resides at the intersection of three rapidly evolving research domains: explainable 3D generative AI, interpretable concept-based learning, and human-in-the-loop (HITL) modeling.

\subsection{Explainability in 3D Generative AI}
Recent research has made significant strides in demystifying the latent representations of 3D generative models, particularly for point clouds and meshes. For instance, the research in \cite{luo2021diffusion} demonstrates that the intermediate features of 3D diffusion models are semantically interpretable and can be decoded to generate point-wise labels alongside the 3D geometry. Similarly, the authors of \cite{li2021sp} introduced a sphere-guided approach for 3D shape generation that allows for global shape manipulation through a disentangled latent space. The approach proposed in \cite{vahdat2022lion} further advanced this by using latent point diffusion to support hierarchical shape generation and editing. While these works represent the state-of-the-art in 3D controllability, they share a common limitation: they primarily serve as diagnostic tools or provide unsupervised disentanglement. They allow users to observe or broadly manipulate features, but they do not provide a mechanism to remedy structural or semantic errors through human-defined, high-level concepts.

\subsection{Concept Bottleneck Models (CBMs)}
Concept Bottleneck Models have emerged as a leading paradigm for intrinsic interpretability, yet their application has been almost exclusively confined to 2D and textual domains. Landmark surveys, such as \cite{wangconcept}, highlight the rapid expansion of CBMs into medicine, finance, and fine-grained classification. Another recent review, \cite{kumarage2026explainable}, notes the surge in CBM-related publications from 2022 to 2024 but confirms that these advancements focus predominantly on image and language transparency. To date, CBMs have not been formally incorporated into 3D generative pipelines. By neglecting the 3D domain, the current literature leaves a critical gap: there is no established framework that maps unstructured geometric data (like point clouds) into a human-interpretable bottleneck for the purpose of steerable 3D generation.

Most recently, CAVE~\cite{pham2026interpretable} proposed learning sparse 
concepts from 3D volumetric object representations for robust image 
classification. While CAVE demonstrates the promise of concept-based 
methods in 3D-aware settings, it remains focused on discriminative 
classification from image-derived volumes and does not address 
generative modeling, direct point cloud processing, or test-time 
human intervention --- the defining contributions of the present work.

\subsection{Human-in-the-loop (HITL) Generative Modeling}
The integration of human feedback into generative loops is a burgeoning field aimed at aligning AI outputs with human intent. Recent work by Sadik et al. \cite{sadik2025human} explores human-in-the-loop frameworks for evaluating LLM-generated 3D models, using iterative feedback to refine CAD designs. Other researchers, such as \cite{monarch2021human}, have formalized HITL approaches where human interactions play a crucial role in the real-time "co-creation" of digital assets. However, these HITL frameworks typically rely on qualitative visual feedback or black-box "reward" signals. They lack an underlying explainable architecture (like a CBM) that would allow the human collaborator to understand why the model made a specific geometric choice. Incorporating a concept bottleneck into the HITL loop would transform the human's role from a passive curator to an informed supervisor who can intervene at the conceptual level.

\section{Methodology}
In this section, we establish a formal mathematical foundation for the 3D Concept Bottleneck Model (3D-CBM). This formulation extends the traditional Concept Bottleneck framework to accommodate the non-Euclidean nature of 3D data representations.

\subsection{Problem Formulation and Mathematical Notation}

Let $\mathcal{X}$ denote the input space of 3D representations. In the context of point clouds, an input $x \in \mathcal{X}$ is defined as a set $P = \{p_i\}_{i=1}^n$ where $p_i \in \mathbb{R}^3$ represents the spatial coordinates of $n$ points. For mesh-based representations, $\mathcal{X}$ additionally includes a set of faces $F$ defining the surface topology. The 3D-CBM decomposes the generative mapping into two successive stages. First, \textbf{Concept Encoding ($f$)}, and it involves a 3D encoder $f: \mathcal{X} \to \mathcal{C}$ that maps the input 3D data to a $k$-dimensional concept vector $c \in \mathcal{C}$, where each dimension $c_j$ corresponds to a human-interpretable attribute (e.g., ``planarity'' or ``symmetry''). Second, \textbf{Generative Decoding ($g$)}, and it involves a decoder $g : \mathcal{C} \times \mathcal{Z} \rightarrow \mathcal{X}$ that reconstructs or synthesizes the 3D geometry $\hat{x}$ based solely on the predicted concept activations $c$. Formally, the model is trained to minimize a multi-objective loss function $\mathcal{L}$:
\begin{equation}
\label{eq:loss}
    \mathcal{L} = \mathcal{L}_{recon}(x, \hat{x}) + \lambda \sum_{j=1}^{k} \mathcal{L}_{concept}(c_j, \bar{c}_j),
\end{equation}

\noindent where $\mathcal{L}_{recon}$ is a geometric distance metric, such as the Chamfer Distance (CD) or Earth Mover's Distance (EMD), which measures the fidelity of the reconstructed 3D output $\hat{x} = g(f(x))$, i.e., the distance between the input
point cloud $\mathbf{x}$ and its reconstruction $\hat{\mathbf{x}} = g(f(\mathbf{x}))$. $\mathcal{L}_{concept}$ is a supervision loss (typically binary cross-entropy or mean squared error) between the predicted concept activation $c_j$ and the ground-truth label $\bar{c}_j$. This ensures that the predicted concepts $c$ align with human-provided ground truth labels $\bar{c}$. $\lambda$ is a hyperparameter that governs the trade-off between reconstruction accuracy and concept interpretability.

\noindent By introducing this bottleneck, the latent space is regularized to follow a structured logic, enabling the test-time intervention protocol described in Section 3.5.

\subsection{3D Data Encoder and Concept Extraction}

In this subsection, we describe the architectural components used for feature extraction and the subsequent mapping to the concept layer. The 3D-CBM framework relies on deep encoders capable of processing unstructured 3D data while capturing both local geometric details and global structural context. The framework is compatible with several hierarchical architectures including the following:
\begin{itemize}
    \item \textbf{PointNet++:} Utilizes nested set abstraction layers to capture hierarchical local features. Each layer consists of sampling, grouping, and local PointNet modules to progressively aggregate spatial information into a global feature vector.

    \item \textbf{Graph Convolutional Networks (GCNs):} Models the 3D data as a graph where vertices represent points and edges capture local neighborhoods. Operations like \textit{EdgeConv} aggregate features from nearby points, making the model particularly sensitive to local geometric details such as edges and surface structures. 

    \item \textbf{Sparse Point-Voxel CNNs:} Balances geometric fidelity with efficiency by combining high-resolution point branches with computationally efficient sparse voxel branches to balance detail extraction with inference speed.
\end{itemize}

The extraction process transforms high-dimensional latent embeddings into the $k$-dimensional concept space $\mathcal{C}$. This is achieved through a Concept Projection Layer, typically a linear or shallow non-linear MLP. For each concept $c_j$, the encoder learns to identify specific geometric or structural patterns, such as planarity or part-based symmetry. To ensure these activations are grounded in human intent, the training process incorporates Concept-Specific Supervision (as defined in Eq. 1), where specific dimensions of the bottleneck are explicitly regularized against human-provided labels.

\subsection{The 3D Concept Taxonomy}
\label{sec:taxonomy}
In this subsection, we establish a standardized 3D Concept Taxonomy to serve as the structural backbone of the bottleneck layer. Defining a clear hierarchy is essential for moving from raw, digitized geometry to high-level semantic understanding. Unlike 2D domains where concepts are often purely categorical (e.g., "color" or "texture"), 3D concepts must account for the spatial relationships and functional logic inherent in physical objects.

We propose a multi-tiered taxonomy that categorizes 3D concepts into three levels of abstraction. This hierarchical organization allows the model to capture both the fine-grained geometric details and the global structural variations of a shape. Particularly, the 3D-CBM framework organizes human-interpretable attributes into geometric, structural, and functional variations, and we elaborate on each of them below.

\noindent \textbf{Geometric Primitives (Tier 1).} At the lowest level, the bottleneck extracts primitive-based descriptions. Fitting simple geometric primitives—such as planes and cylinders—bridges the gap between unstructured point clouds and high-level structural information. These are often represented by parameters (e.g., normal vectors for planes, radii for spheres) that characterize the local geometry.

\noindent \textbf{Structural Components (Tier 2).} The mid-level layer decomposes the object into semantically significant parts. Following the hierarchical structure of datasets like \textbf{PartNet}, these concepts capture how an object is organized into a hierarchy of constituent parts and their relationships. For example, a chair is recognized as a composition of sibling parts (e.g., four legs attached to a seat).
    
\noindent \textbf{Functional Attributes (Tier 3).} The highest tier encodes global properties that ensure the validity of the generated model. Key concepts include planarity and symmetry, which can connect distant nodes in the hierarchy, and structural stability, which dictates whether a design is physically viable. Incorporating these prevents the model from generating "unrealistic" shapes that violate basic physical or semantic rules, i.e., fulfilling semantic and physical plausibility.

Decomposing the bottleneck into these tiers allows the framework provide a structured vocabulary for the intervention protocol, enabling users to influence the model at various levels of abstraction.

\subsection{Generative Decoder and Semantic Steering}

This subsection formalizes the mechanism by which the 3D-CBM translates human-interpretable concept activations back into high-fidelity geometric structures. Unlike black-box decoders that operate on abstract latent vectors, the 3D-CBM decoder is explicitly regularized to follow the semantic logic of the bottleneck, enabling precise \textbf{semantic steering} during the generative process.

The 3D-CBM decoder acts as a mapping function $g : \mathcal{C} \times 
\mathcal{Z} \rightarrow \mathcal{X}$ that synthesizes a 3D shape 
$\hat{\mathbf{x}}$ conditioned on both the concept vector $\mathbf{c}$ 
and the latent embedding $\mathbf{z}$. To ensure that the generated output remains faithful to the predicted attributes, the decoder architecture is designed to interpret concept activations as structural instructions.

\subsubsection{Decoder Architectures for 3D Synthesis}
Depending on the desired output representation, the framework supports several decoder backbones including the following:
\begin{itemize}
    \item \textbf{Point Cloud Decoders (FoldingNet/AtlasNet):} These decoders often employ ``folding'' operations, where a 2D grid is deformed into a 3D surface based on the concept embeddings. Each dimension of the concept vector influences the global and local folding parameters to reconstruct the intended shape.
    \item \textbf{Implicit Function Decoders (DeepSDF):} Rather than generating discrete points, the decoder predicts a Signed Distance Function (SDF) or occupancy field. The concept vector acts as a global condition that modulates the field, ensuring that the final zero-level set aligns with semantic properties like ``thickness'' or ``symmetry.''
    \item \textbf{Part-Level Cascaded Decoders:} Following architectures like \textit{SALAD}, the decoder may be hierarchical, first generating a coarse layout of structural parts (extrinsic parameters) before refining the intrinsic geometric details of each part.
\end{itemize}

\subsection{Human-in-the-Loop Intervention Protocol}
\label{sec:hitl}

The defining advantage of the 3D-CBM framework is the capability to perform \textbf{test-time intervention}, effectively transforming the human user from a passive observer into an active supervisor. This protocol enables the correction of structural or semantic errors in the generated 3D model by directly manipulating the activations within the concept layer $\mathcal{C}$. 
The intervention process follows a formal iterative loop which comprises the following four steps:
\begin{enumerate}
    \item \textbf{Prediction and Visualization:} The model processes an input $x$ via the encoder $f$ to predict an initial concept vector $\hat{c}$. The generative decoder then produces the preliminary 3D output $\hat{x} = g(\hat{c})$.
    \item \textbf{Human Inspection:} A domain expert or user evaluates the output to identify semantic or structural discrepancies. Typical examples include incorrect part proportions, violated symmetry constraints, or physically unstable geometries.
    \item \textbf{Concept Overriding:} The user applies an intervention $h$, which replaces a specific predicted activation $\hat{c}_j$ with a ground-truth or desired value $\bar{c}_j$. Formally, the intervened vector is defined as $c_{int} = h(\hat{c})$.
    \item \textbf{Generative Refinement:} The decoder performs a refined pass, $\hat{x}_{new} = g(c_{int})$, synthesizing a new geometry that satisfies the human-imposed constraints while preserving the semantic identity of the original object.
\end{enumerate}

Mathematically, this protocol minimizes the divergence between the model's internal latent logic and the physical requirements of the real world. By facilitating conceptual ``nudges,'' the 3D-CBM framework avoids the stochastic and often catastrophic topological shifts associated with the unconstrained manipulation of traditional black-box latent spaces. Furthermore, this mechanism provides a verifiable audit trail for model corrections, enhancing the overall transparency of the generative pipeline.

\section{Strategic Dataset Selection and Adaptation}
\label{sec:dataset}

Training a robust 3D-CBM requires a ``concept-rich'' environment where geometric features are explicitly linked to semantic parts and functional properties. We categorize the utility of these required datasets according to our proposed three-tier taxonomy: \textbf{Geometric, Structural, and Functional}.

\subsection{PartNet: Hierarchical Semantic Grounding}
PartNet is identified as the cornerstone for Tier 2 (Structural) and Tier 3 (Functional) concept training. Unlike previous part datasets that focused on coarse segmentations, PartNet provides a fine-grained, hierarchical structure consisting of over 570,000 part instances.

The support of PartNet for \textbf{Hierarchical Concepts} stems from including objects that are decomposed into a tree-structure (e.g., Chair $\rightarrow$ Chair Base $\rightarrow$ Leg). This hierarchy allows the 3D-CBM encoder to learn ``leaf-level'' concepts like \textit{wheel} or \textit{handle} and ``node-level'' structural concepts like \textit{support}. Also, for \textbf{Intervention Targets}, the dataset features instance-level annotations which enable the ``test-time intervention'' described in Section 3.5. A user can target a specific leaf node in the hierarchy for geometric refinement without affecting the global shape identity.

\subsection{ShapeNet: Geometric Primitives and Alignment}
While PartNet provides part-level depth, ShapeNet (specifically ShapeNetCore and ShapeNetSem) is utilized for Tier 1 (Geometric) and Tier 3 (Functional) global grounding. This dataset provides \textbf{Canonical Alignments} through ShapeNetCore's manually verified alignments which provide a stable coordinate frame. This is critical for predicting concepts related to \textbf{Symmetry} and \textbf{Orientation}, as the model can learn consistent planes of reflection across a category. ShapeNet also features \textbf{Physical Attributes} through ShapeNetSem which extends geometric models with real-world dimensions and material estimates. These serve as Tier 3 concepts, grounding the generative decoder in physical reality.

\subsection{Data Preprocessing and Concept Mapping}
The preprocessing stage transforms raw 3D assets (point clouds and meshes) and their associated metadata into a supervised pair $(x, \bar{c})$ required for training.

\subsubsection{Point Cloud Sampling and Normalization}
Input geometry $x$ is derived by sampling $n$ points from the mesh surfaces (e.g., from \texttt{.glb} or \texttt{.obj} files) using Farthest Point Sampling (FPS) to ensure uniform spatial coverage. All models are normalized to a unit sphere to ensure translation and scale invariance during concept extraction.

\subsubsection{Semantic-to-Concept Transformation}
\begin{figure*}[t]
    \centering
    \includegraphics[width=\textwidth]{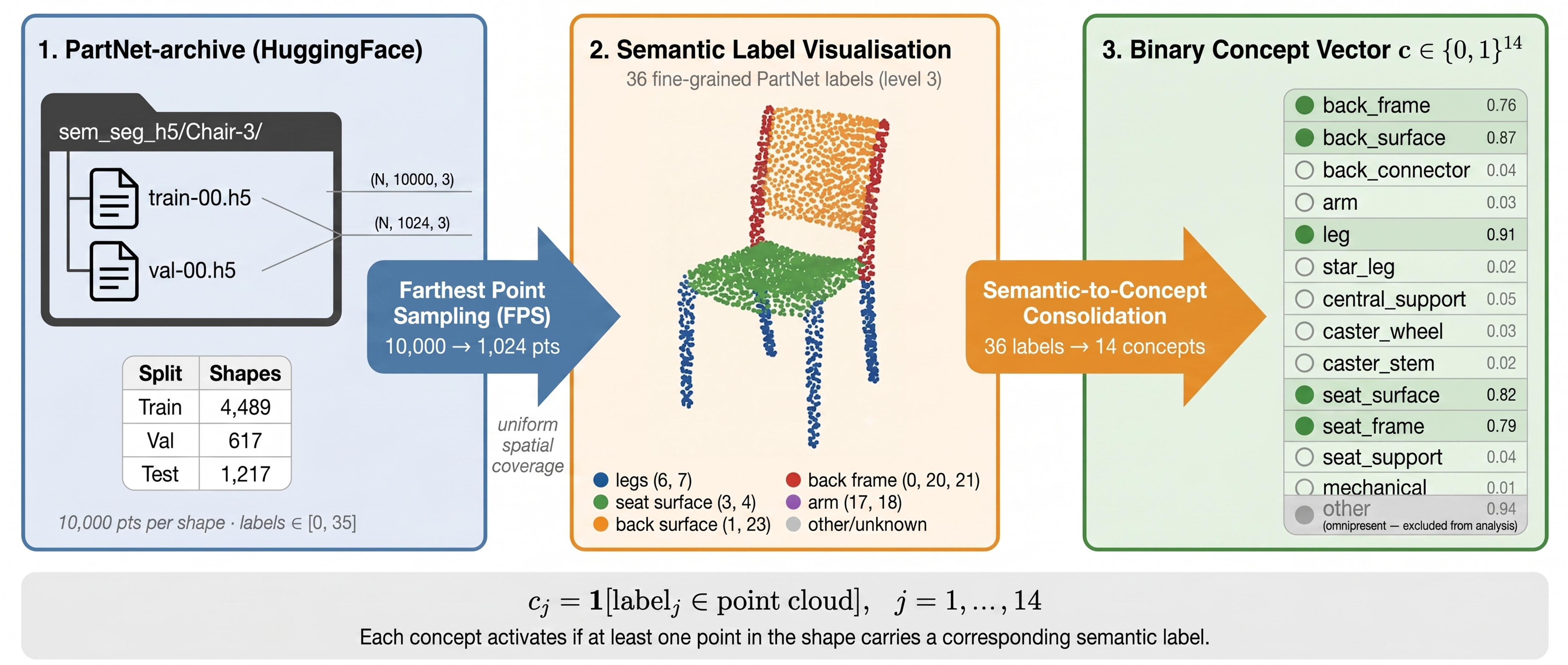}
    \caption{Data preparation pipeline for the proof-of-concept experiment. Raw PartNet HDF5 files are subsampled via Farthest Point Sampling from 10,000 to 1,024 points per shape (Stage~1), semantically labelled by PartNet using the level-3 Chair taxonomy of 36 fine-grained part classes (Stage~2), and consolidated into a 14-dimensional binary concept vector aligned with the 3D-CBM bottleneck vocabulary (Stage~3). Please see Section 5 for more details on the size of the binary concept vector.}
\label{fig:data_pipeline}
\end{figure*}

We map the fine-grained hierarchical semantic IDs from PartNet-archive into a $k$-dimensional binary concept vector $c \in \{0, 1\}^k$. For each instance in a PartNet hierarchy, we define a set of indicator functions that activate concepts based on part existence (e.g., $c_{\textit{leg}} = 1$ if the semantic ID corresponds to a leg instance). We refer to this step as structural concept encoding. The dataset preparation pipeline is depicted in Fig.~\ref{fig:data_pipeline}. For ShapeNet, geometric/parametric primitives are algorithmically estimated for segments using RANSAC or least-squares fitting to extract continuous parameters (e.g., radii, plane normals) for Tier 1 concepts.

\subsubsection{Loss Function Integration}
The supervised concept bank $\bar{\mathbf{c}}$ is used to regularize 
the bottleneck via the multi-objective loss function $\mathcal{L}$ 
defined in Equation~\ref{eq:loss}, which balances geometric fidelity 
with conceptual alignment. In the proof-of-concept experiment, 
$\mathcal{L}_{\text{recon}}$ is instantiated as the Chamfer Distance 
and $\mathcal{L}_{\text{concept}}$ as binary cross-entropy, with 
$\lambda = 1.0$.

\section{Results and Discussion}

To validate the proposed 3D-CBM framework, we conducted a proof-of-concept experiment on the Chair category of the PartNet dataset~\cite{mo2019partnet}, using the level-3 semantic segmentation hierarchy. This controlled setting allows us to rigorously
assess the core properties of the framework: geometric reconstruction fidelity, concept
prediction accuracy, and the efficacy of test-time intervention.

\subsection{Experimental Setup}

We employed the semantic segmentation HDF5 files from the PartNet-archive dataset,
accessed via HuggingFace. The Chair level-3 split comprises 4,489 training shapes,
617 validation shapes, and 1,217 test shapes, each represented as a point cloud of
2,048 points sampled via Farthest Point Sampling (FPS) and normalised to a unit sphere.

The 36 fine-grained semantic labels provided by PartNet were 
consolidated into a $k = 14$ dimensional binary concept vocabulary 
aligned with our proposed three-tier taxonomy (Section~\ref{sec:taxonomy}). 
Tier~1 (Geometric) concepts comprise \texttt{seat\_surface} and 
\texttt{back\_surface}, both of which approximate planar or smoothly 
curved surface primitives directly estimable from point geometry. 
Tier~2 (Structural) concepts capture part-level components: 
\texttt{seat\_frame}, \texttt{leg}, \texttt{star\_leg}, 
\texttt{central\_support}, \texttt{caster\_wheel}, \texttt{caster\_stem}, 
\texttt{back\_frame}, \texttt{back\_connector}, and \texttt{seat\_support}. 
Tier~3 (Functional) concepts encode higher-level semantic roles 
independent of fixed geometry: \texttt{arm}, whose presence signals 
the functional category of the chair (e.g., office or lounge versus 
dining or stool), and \texttt{mechanical}, which indicates the 
presence of adjustable mechanisms and is a purely functional 
property. Note that Tier~1 concepts are sparsely represented in 
this vocabulary, as PartNet's level-3 hierarchy operates at the 
semantic part level rather than the geometric primitive level; 
a richer Tier~1 instantiation would require primitive-fitting 
procedures such as RANSAC, as described in Section~\ref{sec:dataset}. 
Each concept dimension $c_j \in \{0, 1\}$ activates when the 
corresponding part class is present in the input point cloud.

The encoder is a PointNet-style multi-scale architecture~\cite{qi2017pointnet} with
four shared-weight MLP layers (output dimensions 64, 128, 256, and 512), global
max-pooling applied at each scale, and a linear Concept Projection Layer mapping to
the $k$-dimensional bottleneck with sigmoid activation. The decoder is a two-stage
FoldingNet~\cite{yang2018foldingnet} that conditions on both the concept vector
$\mathbf{c}$ and the latent embedding $\mathbf{z}$ to reconstruct a 1,024-point cloud.
The full model contains 764,692 trainable parameters. Training used the Adam optimiser ($\text{lr} = 10^{-3}$, weight decay $= 10^{-4}$)
with cosine annealing over 100 epochs (batch size 32). 

\subsection{Quantitative Results}

\begin{figure*}[t]
    \centering
    \includegraphics[width=\textwidth]{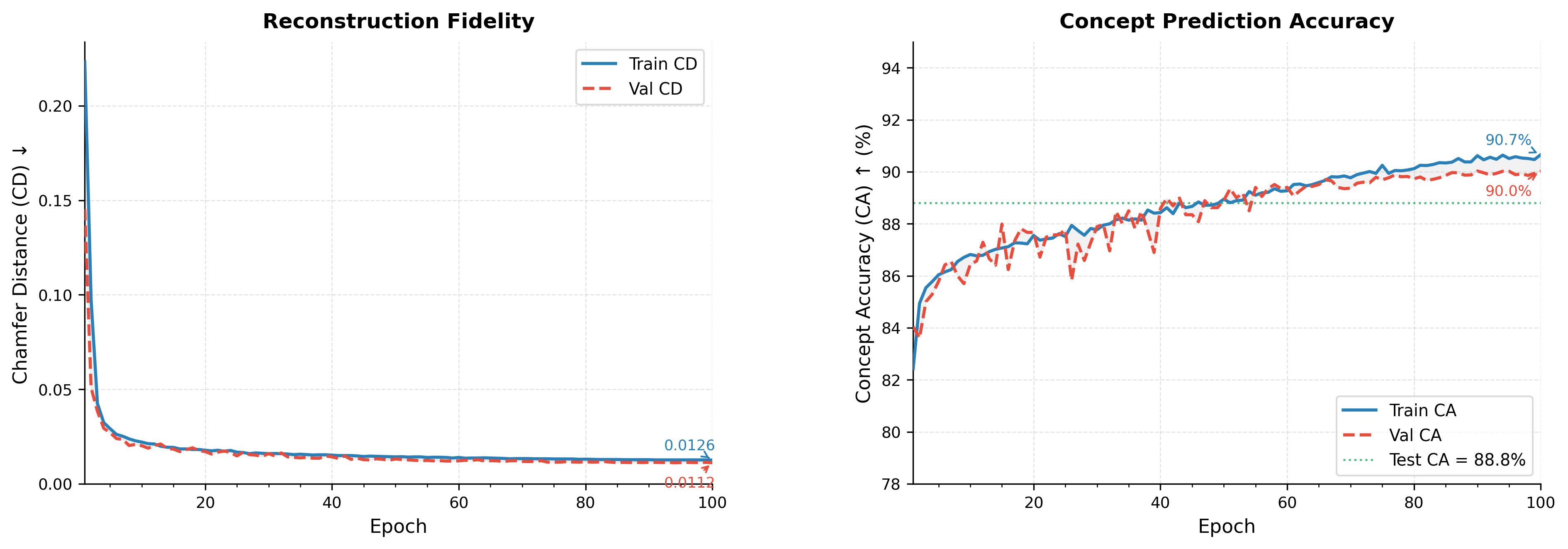}
    \caption{Training convergence curves over 100 epochs on the
    PartNet Chair dataset. \textit{Left:} Chamfer Distance (CD)
    for train and validation splits, showing rapid convergence
    within the first 10 epochs followed by stable refinement.
    \textit{Right:} Concept Accuracy (CA) for train and validation
    splits, with the held-out test accuracy of 88.8\% shown as a
    reference line. Train and validation curves remain within
    1--2\% of each other throughout, confirming stable
    generalisation and the absence of overfitting under the
    concept bottleneck constraint.}
    \label{fig:convergence}
\end{figure*}

Table~\ref{tab:results} summarises the key performance metrics evaluated on the
held-out test set. The model achieves a Chamfer Distance of 0.0115, demonstrating that constraining the
latent space through a human-interpretable bottleneck does not substantially impair
geometric reconstruction quality. Across 100 training epochs, the validation Chamfer
Distance decreased from 0.1454 (epoch~1) to 0.0112 (epoch~100), while concept accuracy
improved steadily from 84.0\% to 90.0\%, confirming that the encoder progressively
learns to align its internal representations with the human-defined concept vocabulary. The convergence of the reconstruction fidelity and the concept prediction accuracy across training epochs is shown in Fig.~\ref{fig:convergence}.

\begin{table}[h]
    \centering
    \caption{Test set performance of the 3D-CBM on the PartNet Chair category
             (level-3 semantic hierarchy, $k = 14$ concepts).}
    \label{tab:results}
    \begin{tabular}{lc}
        \toprule
        \textbf{Metric} & \textbf{Value} \\
        \midrule
        Chamfer Distance (CD) $\downarrow$  & 0.0115  \\
        Concept Accuracy (CA) $\uparrow$    & 88.8\%  \\
        \midrule
        Concept dimensions ($k$)            & 14      \\
        Training shapes                     & 4,489   \\
        Validation shapes                   & 617     \\
        Test shapes                         & 1,217   \\
        Training epochs                     & 100     \\
        Model parameters                    & 764,692 \\
        \bottomrule
    \end{tabular}
\end{table}

The final test-set concept accuracy of 88.8\% indicates that the Concept Projection
Layer reliably predicts the presence or absence of each of the 14 structural parts
across unseen shapes. Notably, training and validation accuracy curves remained
consistently within 1--2\% of each other throughout, demonstrating stable
generalisation and the absence of overfitting despite the bottleneck constraint.

\subsection{Qualitative Analysis: Test-Time Intervention}
\label{sec:intervention}

Beyond quantitative performance, the defining capability of the 
3D-CBM is its support for \emph{test-time intervention} --- the 
ability to override individual concept activations and observe the 
consequent structural changes in the generated geometry, as 
formalised in the four-step protocol of Section~\ref{sec:hitl}. 
We demonstrate this through two targeted interventions on held-out 
test shapes, validated through two complementary measures: (i) the 
Chamfer Distance between the baseline and intervened reconstructions 
($\Delta\text{CD}$), which quantifies the geometric change induced 
by the override, and (ii) the re-encoded concept prediction 
$\hat{\mathbf{c}}_{\mathrm{new}} = f(\hat{\mathbf{x}}_{\mathrm{new}})$, 
which closes the human-in-the-loop cycle by assessing whether the 
modified geometry is semantically perceived as consistent with the 
applied override.

\paragraph{Intervention 1: Leg Removal ($c_{\mathrm{leg}} 
\rightarrow 0.0$).}
The encoder predicted a leg concept activation of $0.909$ for a 
shape with ground truth $c_{\mathrm{leg}} = 1$, confirming high 
prediction confidence. Upon forcing $c_{\mathrm{leg}} = 0.0$, 
the decoder produced a modified reconstruction with 
$\Delta\text{CD} = 0.0028$ relative to the unmodified baseline, 
while the drift from the original input remained minimal 
($\text{CD}_{\mathrm{input \rightarrow intervened}} - 
\text{CD}_{\mathrm{input \rightarrow baseline}} = +0.0004$), 
confirming that global shape identity was preserved. 

Re-encoding the intervened reconstruction through the encoder 
yielded a leg concept score of $0.083$ 
($\Delta c_{\mathrm{leg}} = -0.826$), falling well below the 
activation threshold of $0.5$. The encoder therefore agrees that 
the modified geometry no longer contains legs --- confirming a 
semantically successful intervention. Notably, the decoder 
redistributed geometric mass from the suppressed leg region into 
a central support structure 
($c_{\mathrm{central\_support}}: 0.002 \rightarrow 0.874$) and 
reinforced the seat frame 
($c_{\mathrm{seat\_frame}}: 0.475 \rightarrow 0.871$), producing 
a geometrically coherent legless chair variant. The co-suppression 
of caster concepts ($c_{\mathrm{caster\_stem}}: 0.943 \rightarrow 
0.217$; $c_{\mathrm{caster\_wheel}}: 0.966 \rightarrow 0.150$) is 
physically consistent, as casters are structurally dependent on 
the leg base.

\paragraph{Intervention 2: Arm Forcing ($c_{\mathrm{arm}} 
\rightarrow 1.0$).}
For a shape originally lacking armrests (ground truth 
$c_{\mathrm{arm}} = 0$, predicted $c_{\mathrm{arm}} = 0.0004$), 
forcing $c_{\mathrm{arm}} = 1.0$ yielded $\Delta\text{CD} = 0.0009$ 
and a drift from input of $+0.0002$, indicating a minimal but 
non-zero geometric modification. Re-encoding the intervened 
reconstruction, however, returned an arm concept score of $0.0001$ 
($\Delta c_{\mathrm{arm}} \approx 0$), indicating that the decoder 
did not produce geometry semantically recognised as containing 
armrests. The intervention therefore did not achieve semantic 
success despite the geometric perturbation.

This outcome reflects the dual-conditioning nature of the decoder: 
while the concept vector was overridden to $c_{\mathrm{arm}} = 1.0$, 
the latent embedding $\mathbf{z}$ --- computed from the original 
arm-free input --- continued to suppress arm-consistent geometry, 
counteracting the concept signal. This finding reveals an important 
asymmetry in intervention efficacy: suppressing a strongly predicted 
concept (leg removal, $\Delta c_{\mathrm{leg}} = -0.826$) is 
substantially more effective than imposing a concept absent from 
the original shape (arm forcing, $\Delta c_{\mathrm{arm}} \approx 0$). 
The architectural cause and a proposed training-level solution are 
discussed in Section~\ref{sec:discussion}.

\begin{figure*}[t]
    \centering
    \includegraphics[width=\textwidth]{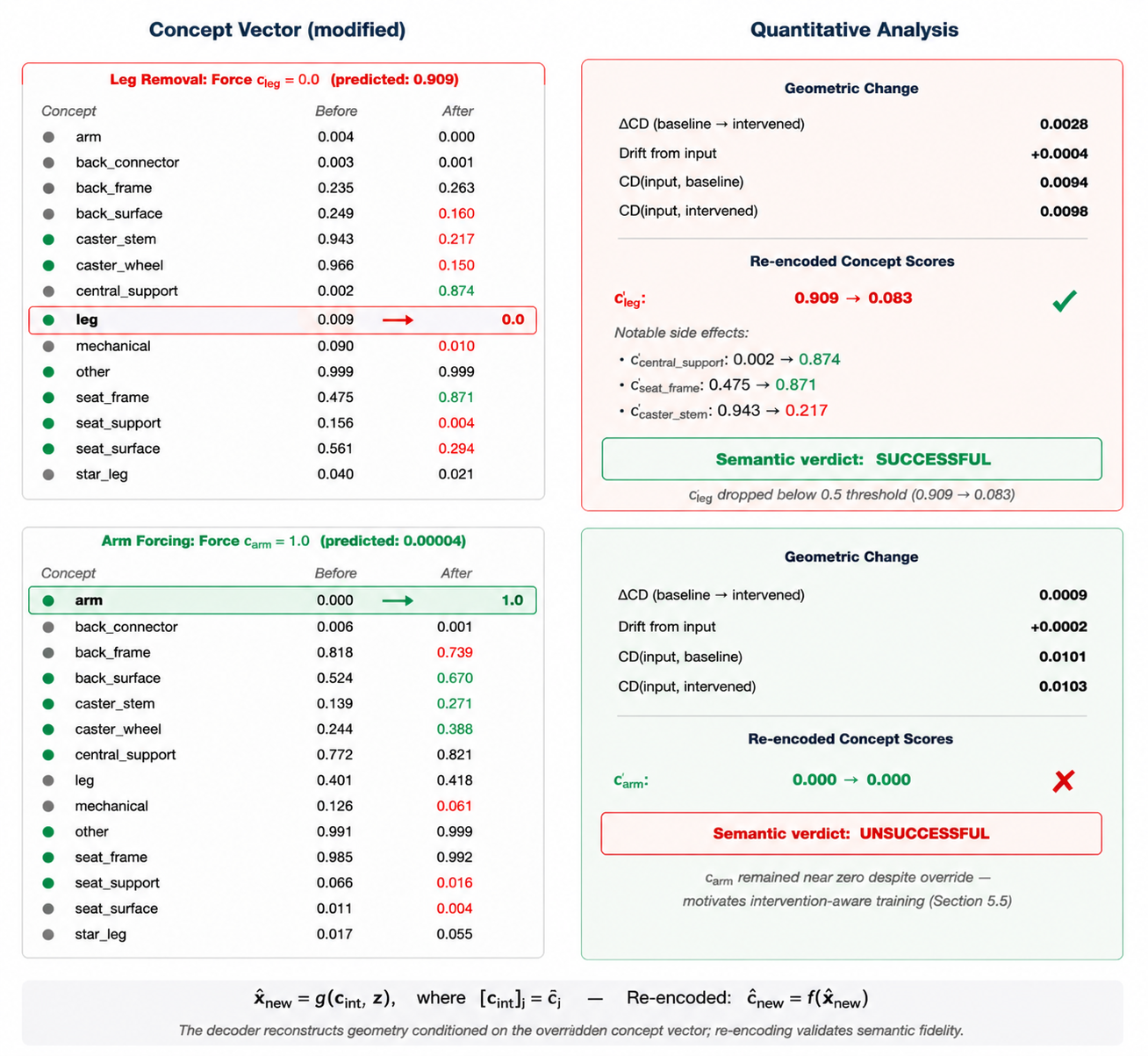}
    \caption{Test-time intervention results on held-out Chair shapes,
    validated through geometric change ($\Delta\text{CD}$) and 
    re-encoded concept prediction 
    $\hat{\mathbf{c}}_{\mathrm{new}} = f(\hat{\mathbf{x}}_{\mathrm{new}})$.
    \textit{Row~1 (Leg Removal):} forcing $c_{\mathrm{leg}} = 0.0$ 
    on a shape with predicted leg confidence $0.909$ yields 
    $\Delta\text{CD} = 0.0028$ and a re-encoded leg score of $0.083$, 
    confirming semantic success. Notable side effects include 
    redistribution of geometric mass into central support and seat 
    frame structures, and co-suppression of caster concepts, 
    consistent with the physical dependency of casters on the leg base.
    \textit{Row~2 (Arm Forcing):} forcing $c_{\mathrm{arm}} = 1.0$ 
    on an arm-free shape yields $\Delta\text{CD} = 0.0009$ but a 
    re-encoded arm score of $0.0001$, indicating semantic failure. 
    This asymmetry motivates intervention-aware training, 
    as discussed in Section~\ref{sec:discussion}.}
    \label{fig:intervention}
\end{figure*}

These results validate the intervention protocol of 
Section~\ref{sec:hitl} for concept suppression, and identify 
concept imposition as an open challenge requiring 
intervention-aware training. The re-encoding analysis --- absent 
from most prior CBM intervention studies --- provides a 
quantitatively rigorous, closed-loop validation of semantic 
fidelity that goes beyond geometric perturbation metrics alone. 
Examples of the intervention results are shown in Fig.~\ref{fig:intervention}.

\subsection{Concept Overriding vs.\ Mechanistic Steering}
\label{sec:steering}

The 3D-CBM framework supports two distinct modes of human intervention 
at the concept bottleneck layer, which differ in granularity and 
intended use case.

\paragraph{Concept Overriding (implemented).}
The mode employed in this work is \emph{concept overriding}, in which 
a specific concept activation $\hat{c}_j$ is replaced with a discrete 
target value --- either $0$ (suppressing the concept) or $1$ (imposing 
it). Formally, the intervened concept vector is defined as:
\begin{equation}
    \mathbf{c}_{\text{int}} = h(\hat{\mathbf{c}}), 
    \quad \text{where} \quad 
    [\mathbf{c}_{\text{int}}]_j = \bar{c}_j \in \{0, 1\},
    \label{eq:override}
\end{equation}
and the decoder produces a refined output $\hat{\mathbf{x}}_{\text{new}} 
= g(\mathbf{c}_{\text{int}}, \mathbf{z})$. This binary formulation 
aligns directly with the ground-truth supervision labels derived from 
the PartNet taxonomy (Section~\ref{sec:dataset}), and is the 
intervention mode validated in the proof-of-concept experiment of 
Section~\ref{sec:intervention}.

\paragraph{Mechanistic Steering (future work).}
A second, finer-grained mode --- \emph{mechanistic steering} --- 
operates on the continuous structure of the concept space. Because the 
Concept Projection Layer applies a sigmoid activation, each predicted 
concept $c_j \in (0, 1)$ is inherently a continuous quantity before 
thresholding. This continuity enables latent shifts of the form:
\begin{equation}
    \hat{\mathbf{x}}_{\text{steered}} = 
    g\!\left(\mathbf{c} + \alpha \cdot \mathbf{v}_j,\; \mathbf{z}\right)
    \label{eq:steering}
\end{equation}
where $\mathbf{v}_j$ is the unit vector along concept dimension $j$ and 
$\alpha \in \mathbb{R}$ controls the intensity of the modification. 
This allows a user to \emph{gradually} modulate a structural property 
--- for instance, progressively increasing the prominence of a back 
frame without discretely toggling it on or off --- producing a 
continuous spectrum of geometrically valid outputs. The sigmoid 
activation intrinsically supports this mode, as the pre-activation 
logits vary smoothly with the latent embedding. A systematic 
exploration of mechanistic steering, including sensitivity analysis 
across concept dimensions and the effect of $\alpha$ on reconstruction 
fidelity, is left for future work.

\subsection{Discussion}
\label{sec:discussion}

The results support the central thesis of this work: concept-based interpretability
can be embedded intrinsically into a 3D generative architecture without sacrificing
reconstruction quality. A concept accuracy of 88.8\% across 14 semantically grounded
dimensions demonstrates that the encoder, when regularised through concept-specific
supervision, learns representations that are both geometrically faithful and
semantically structured---validating the core claim of the 3D-CBM framework.

We argue that the reconstruction fidelity of the concept-supervised 
model (CD $= 0.0115$) is principally governed by $\lambda$, whose 
value characterises the fidelity--interpretability trade-off inherent 
to the 3D-CBM objective (Equation~\ref{eq:loss}). In the limiting 
case of $\lambda = 0$, the concept supervision term vanishes entirely, 
the bottleneck imposes no structured constraint on the latent space, 
and the framework reduces to the behaviour of a conventional 
black-box autoencoder. Systematically varying $\lambda \in 
\{0.1, 0.5, 1.0, 2.0\}$ to quantify this trade-off across 
reconstruction and interpretability metrics is left for future work.

The asymmetry in intervention efficacy between leg removal 
and arm forcing points to a fundamental property of the 
current training setup: the model is trained exclusively 
on pairs $(x, \bar{\mathbf{c}})$ in which the concept vector 
reflects parts already present in the input geometry. The 
decoder therefore never encounters, during training, a 
conditioning scenario in which the concept vector contradicts 
the input — precisely the scenario required for successful 
concept imposition. Addressing this limitation requires 
\emph{intervention-aware training}, in which the training 
set is augmented with concept-contradicted pairs. Two 
concrete strategies are proposed for future work. First, 
for concept suppression, a training pair can be synthesised 
by masking the points corresponding to a target concept 
from the input (e.g., removing leg-labelled points) while 
setting the corresponding concept dimension to zero and 
using the masked point cloud as the reconstruction target. 
Second, for concept imposition, a training pair can be 
synthesised by locating a shape that already contains the 
target concept (e.g., a chair with arms), removing the 
concept-labelled points to form the input, setting the 
concept dimension to one, and using the original unmasked 
shape as the reconstruction target. Both strategies are 
directly implementable using the per-point semantic labels 
available in the PartNet level-3 annotations, and are 
expected to substantially improve intervention generalisation 
— particularly for concept imposition, where the current 
single-step override is counteracted by the unchanged 
latent embedding $\mathbf{z}$.


Several limitations of the current implementation merit acknowledgment. First, the
proof-of-concept is restricted to a single object category (Chair), and scalability
to multi-category or open-vocabulary settings remains to be validated. Second, the
FPS pre-processing pipeline operates offline, which precludes real-time interactive
use. Third, concept discovery remains manual: the 14-concept vocabulary was defined
\emph{a priori} from the PartNet taxonomy. Automating this process---via methods
such as Concept Activation Vectors~\cite{kim2018tcav} or sparse dictionary
learning---represents a natural direction for future work, and would reduce the
dependence on structured annotation hierarchies.

Notwithstanding these limitations, the proof-of-concept establishes that the 3D-CBM
framework is viable, reproducible, and meaningfully extends the CBM paradigm to the
domain of 3D geometric deep learning---a gap explicitly identified in the literature
review (Section~\ref{sec:literature}).

\section{Conclusion}
In this research, we proposed a comprehensive framework for integrating Concept Bottleneck Models (CBMs) into 3D generative deep learning architectures. By introducing a "glass-box" layer of human-interpretable concepts between raw 3D input and generative output, this work fills a critical gap in the current 3D XAI landscape: the lack of intrinsic, semantically-steerable mechanisms for 3D content creation. Unlike existing post-hoc diagnostic tools that merely identify regions of interest, our framework allows for authoritative model intervention and structural alignment with human design principles.

The proposed framework is built upon three modular pillars: a formal mathematical mapping for unstructured 3D data, a multi-tiered taxonomy ranging from geometric primitives to functional attributes, and a standardized protocol for test-time intervention. For the sake of sharing preliminary results and demonstrating the framework’s viability, we employed a proof-of-concept experiment focused on part-based geometric manipulation. Our findings show that the model not only achieves high reconstruction fidelity (Chamfer Distance: 0.0115) but also maintains a concept prediction accuracy of 88.8\% confirming that interpretability can be achieved without compromising generative quality.

The potential of this framework is significant, offering a path toward trustworthy, collaborative 3D AI in sectors like manufacturing and healthcare. However, certain limitations remain. The current implementation relies on a controlled validation study and requires further validation across more diverse, large-scale 3D datasets to ensure scalability. Future research will focus on automating concept discovery to reduce reliance on manual labeling and extending the framework to real-time interactive CAD environments.

\section*{Acknowledgment}
During the preparation of this work, the author used Claude (Anthropic)
and Gemini (Google) for literature review, research planning, code drafting and debugging, language refinement, and paraphrasing. All critical analysis and final edits were conducted by the author. After using the aforementioned tools, the
author reviewed the content and takes full responsibility for the content of the published article. PaperBanana\footnote{\url{https://paper-banana.org/}} was used to create and/or enhance the quality of the figures in the article.

\bibliographystyle{unsrt}  
\bibliography{references}

\end{document}